\documentclass[letterpaper, 10 pt, conference]{ieeeconf}  

\IEEEoverridecommandlockouts                              

\usepackage[pdftex]{graphicx}
\usepackage{stix}
\usepackage[cmex10]{amsmath}
\usepackage{bm}
\usepackage{algorithm,algpseudocode}
\usepackage{lipsum}
\usepackage{url}
\usepackage{mdwtab}
\usepackage[caption=false]{subfig}
\usepackage{romannum}
\usepackage{threeparttable}

\makeatletter
\newenvironment{breakablealgorithm}
  {
   \begin{center}
     \refstepcounter{algorithm}
     \hrule height.8pt depth0pt \kern2pt
     \renewcommand{\caption}[2][\relax]{
       {\raggedright\textbf{\ALG@name~\thealgorithm} ##2\par}%
       \ifx\relax##1\relax 
         \addcontentsline{loa}{algorithm}{\protect\numberline{\thealgorithm}##2}%
       \else 
         \addcontentsline{loa}{algorithm}{\protect\numberline{\thealgorithm}##1}%
       \fi
       \kern2pt\hrule\kern2pt
     }
  }{
     \kern2pt\hrule\relax
   \end{center}
  }
 \makeatother
\algnewcommand\algorithmicinput{\textbf{Input:}}
\algnewcommand\Input{\item[\algorithmicinput]}
\algnewcommand\algorithmicoutput{\textbf{Output:}}
\algnewcommand\Output{\item[\algorithmicoutput]}
\algnewcommand\algorithmicParameter{\textbf{Parameters:}}
\algnewcommand\Parameters{\item[\algorithmicParameter]}

\algnewcommand\Or{\textbf{or} }

\title{\LARGE \bf
Prediction of Vessel Arrival Time to Pilotage Area Using Multi-Data Fusion and Deep Learning
}

\author{Xiaocai Zhang$^{1}$, Xiuju Fu$^{1}$, Zhe Xiao$^{1\star}$, Haiyan Xu$^{1}$, Xiaoyang Wei$^{1}$, Jimmy Koh$^{2}$, Daichi Ogawa$^{3}$ \\ and Zheng Qin$^{1}$
\thanks{*Corresponding author: Zhe Xiao.}
\thanks{$^{1}$Xiaocai Zhang, Xiuju Fu, Zhe Xiao, Haiyan Xu, Xiaoyang Wei and Zheng Qin are with the Institute of High Performance Computing (IHPC), Agency for Science, Technology and Research (A*STAR), 1 Fusionopolis Way, \#16-16 Connexis, Singapore 138632, Republic of Singapore
        {\tt\small zhang\_xiaocai@ihpc.a-star.edu.sg; xiaoz@ihpc.a-star.edu.sg}}%
\thanks{$^{2}$Jimmy Koh is with the Pilotage and Digital Transformation Department, PSA Marine (Pte) Ltd, 70 West Coast Ferry Road, Singapore 126800, Republic of Singapore}%
\thanks{$^{3}$Daichi Ogawa is with the MTI Co., Ltd. Singapore Branch, 1 Harbourfront Place \#14-01, Harbourfront Tower One, Singapore 098633, Republic of Singapore}%
}

\begin{document}

\maketitle
\thispagestyle{empty}
\pagestyle{empty}

\begin{abstract}
This paper investigates the prediction of vessels' arrival time to the pilotage area using multi-data fusion and deep learning approaches.
Firstly, the vessel arrival contour is extracted based on Multivariate Kernel Density Estimation (MKDE) and clustering.
Secondly, multiple data sources, including Automatic Identification System (AIS), pilotage booking information, and meteorological data, are fused before latent feature extraction.
Thirdly, a Temporal Convolutional Network (TCN) framework that incorporates a residual mechanism is constructed to learn the hidden arrival patterns of the vessels.
Extensive tests on two real-world data sets from Singapore have been conducted and the following promising results have been obtained: 1) fusion of pilotage booking information and meteorological data improves the prediction accuracy, with pilotage booking information having a more significant impact; 2) using discrete embedding for the meteorological data performs better than using continuous embedding; 3) the TCN outperforms the state-of-the-art baseline methods in regression tasks, exhibiting Mean Absolute Error (MAE) ranging from 4.58 min to 4.86 min; and 4) approximately 89.41\% to 90.61\% of the absolute prediction residuals fall within a time frame of 10 min.

\end{abstract}

\section{INTRODUCTION}
Pilotage is a mandatory requirement for many vessels entering and exiting ports worldwide and is an essential component of safe and efficient navigation \cite{hutabarat2020use}. Pilots are highly trained professionals aimed to guide vessel captains, ensuring vessels' safe passage in and out of a port. The pilotage area is a fixed water zone the pilot embarks on the vessel to commence pilotage service.
The efficient operation of ports relies on a delicate balance of timing and coordination, and the Just-In-Time (JIT) arrival of vessels to pilot boarding areas is a critical component of this process. In order to facilitate JIT operations between pilotage service and arriving vessels, it is essential for the pilot operation center to accurately monitor vessel arrival times and adjust schedules accordingly. This not only has the potential to improve the overall efficiency of operations, but it also mitigates the risks associated with vessels waiting at pilot boarding areas as a result of arriving too early.
Given the diverse arrival behaviors that vessels exhibit \cite{zhang2022vessel}, accurately predicting their arrival patterns becomes challenging due to the increased uncertainty. This uncertainty, such as movement behaviors and pilotage operation behaviors, makes it difficult to obtain robust predictions that can withstand potential disruptions or variations in the vessels' schedules.

In this study, we present an approach for predicting vessel arrival time to pilotage areas using multi-data fusion and deep learning.
This method has the potential to enhance the efficiency of port operations by enabling Just-in-Time (JIT) operations at pilot boarding areas. The following are the key contributions to highlight:
\begin{itemize}
\item To the best of our knowledge, this is the first study to explore the forecasting of the arrival time of vessels to pilotage areas;
\item We propose a hybrid approach based on statistical learning and machine learning to identify vessel arrival patterns by extracting vessel arrival contour using historical trajectory data;
\item We first develop a Temporal Convolutional Network (TCN) framework that integrates residual blocks to model the problem of vessel arrival time prediction;
\item We evaluate the performance of our proposed method on data sets obtained from Singapore and compare it with other advanced deep learning algorithms to validate its effectiveness and superiority.
\end{itemize}

The subsequent parts of this article are organized as follows. Section \Romannum{2} presents a summary of previous studies on predicting ship arrival time. In Section \Romannum{3}, we explain the proposed method in detail, including vessel arrival contour extraction and TCN-based model using multi-data fusion. Section \Romannum{4} details the results of the experiments and analyses. Finally, in Section \Romannum{5}, we conclude this study.

\section{RELATED WORK}
The existing approaches for modelling vessel arrival time prediction can be categorized into two types: statistical models and machine learning models. 
The authors in \cite{alessandrini2018estimated} formulated a data-driven path-finding algorithm for vessel arrival time prediction. The algorithm involves a set of parameters conveniently optimized for the area under investigation.
The authors in \cite{park2021vessel} introduced a path-finding algorithm to find possible vessel trajectories using Automatic Identification System (AIS) data. Then the Markov Chain property and Bayesian Sampling were introduced to estimate the Speed Over Ground (SOG) of a vessel. Finally, the Estimated Time of Arrival (ETA) was estimated from the derived SOG values for the predicted trajectory.
Bodunov \textit{et al.} \cite{bodunov2018real} constructed a feed-forward neural network to forecast the arrival time of vessels.
Ogura \textit{et al.} \cite{ogura2021prediction} extracted the route and used Bayesian learning to calculate the speed along the voyage by considering the impact of weather.
In the publication referenced as \cite{salleh2017predicting}, the authors employed a Fuzzy Rule-Based Bayesian Network (FRBBN) to evaluate and forecast the on-time arrival of a liner container vessel at ports of call, even when faced with uncertain circumstances.
In \cite{noman2021towards}, the authors concluded that Gated Recurrent Unit (GRU) outperforms Gradient Boosting Decision Trees (GBDT) and Multi-Layer Perceptron (MLP) based on AIS data.
Yu \textit{et al.} \cite{yu2018ship} addressed the problem of ship arrival prediction (delay or advance arrival prediction) by using three methods, i.e., BPNN, Classification And Regression Tree (CART), and Random Forecast (RF). The prediction results show that the RF model outperforms BP and CART models.
In \cite{balster2020eta}, the authors provided a supervised learning approach for ETA prediction for intermodal freight transport networks involving multiple modes of transportation.
Mekkaoui \textit{et al.} \cite{el2023deep} showed a Long Short-Term Memory (LSTM)-based model for vessel arrival time prediction using different data sources, including vessel characteristics, AIS data, and weather data.
Valero \textit{et al.} \cite{valero2022prediction} conducted a practical analysis of prior studies on predicting ETA for container vessels and determined that RF achieved the best performance. Meanwhile, Cammin \textit{et al.} \cite{cammin2020applications} utilized the name and shipping line type of vessels as features to forecast arrival delays using the Support Vector Regression.

\section{METHOD}
\begin{figure}[htbp]
\begin{center}
\includegraphics[width=0.485\textwidth]{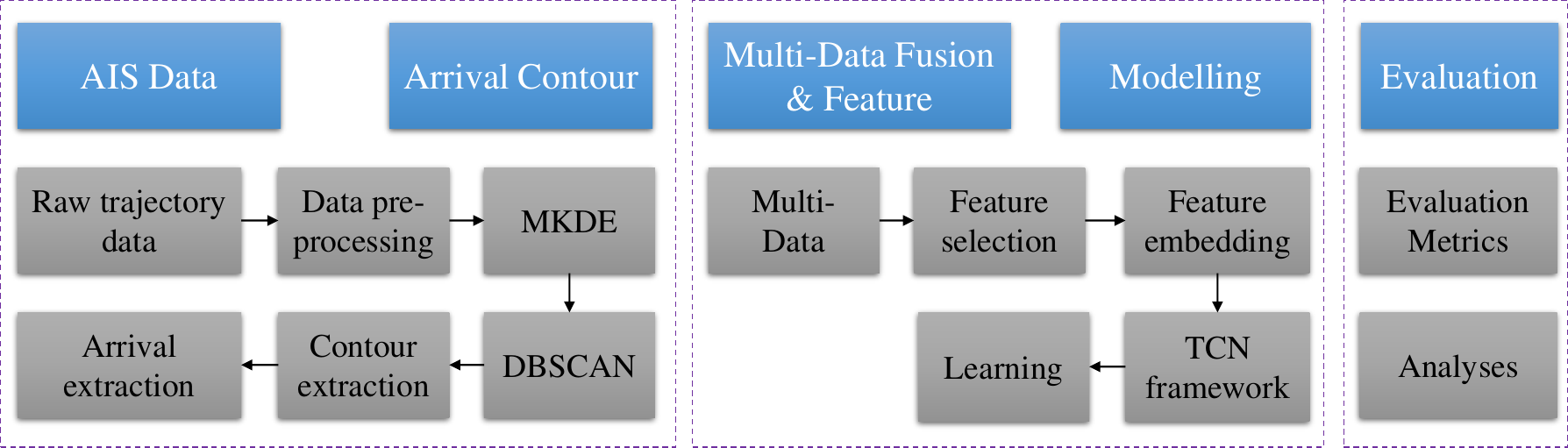}
\caption{The framework of prediction of the vessel arrival time to pilotage area.}
\label{framework}
\end{center}
\end{figure}

The vessel's arrival time is determined as the moment when the vessel reaches the designated pilot boarding areas of the port.
Fig. \ref{framework} illustrates the framework for predicting the arrival time of vessels to the pilotage area. This framework consists of five modules: 1) AIS data, 2) arrival contour, 3) multi-data fusion and feature, 4) modelling, and 5) evaluation. The subsequent sections of this document delve into the specifics of each module. The first two modules, i.e., AIS data and arrival contour, are further explained in Section \ref{cont}, while the subsequent two modules, i.e., multi-data fusion and feature, and modelling, are elaborated in Section \ref{tcn}.

\subsection{Vessel Arrival Contour Extraction}
\label{cont}
Assume that there are a total of $n$ historical trajectories that lead to a Pilot Boarding Ground (PBG), for each AIS trajectory $i$, we can determine its nearest point to the PBG and denote it as $X_{i}$, as shown by
\begin{equation}\label{eq_1}
X = \left ( X_{1},X_{2},\cdots ,X_{n} \right ),
\end{equation}
and
\begin{equation}\label{eq_2}
X_{i}=\left ( x_{i1},x_{i2} \right )^{\textup{T}}\in \mathbb{R}^{2},
\end{equation}
where $x_{i1}$ and $x_{i2}$ are the longitude and latitude, respectively.
The following objective is to estimate the probability density function, represented by $f\left ( x_{1},x_{2} \right )$, of these points $X$.
We propose to use the Multivariate Kernel Density Estimation (MKDE) method to derive an estimate of $f\left ( x_{1},x_{2} \right )$. The formulation of MKDE is expressed as
\begin{equation}\label{eq_3}
\hat{f}_{h}\left ( x_{1},\cdots ,x_{d} \right ) =\frac{1}{n}\sum_{i=1}^{n}\frac{1}{h_{1}\cdots h_{d}}K\left ( \frac{x_{1}-x_{i1}}{h_{1}},\cdots ,\frac{x_{d}-x_{id}}{h_{d}} \right ),
\end{equation}
where $d$ is the dimension of the variables, $K$ is the kernel function, and $h$ is the bandwidth vector. In this study, $d=2$, and the Gaussian kernel is chosen as the kernel.

Following that, those points with high probability density values, e.g., greater than a threshold, are selected for the follow-up step to extract arrival contour and to define the arrival of vessels to the pilot boarding ground, as shown by
\begin{equation}\label{eq_4}
\hat{X} = \left \{ X_{i},\textup{if}\; \hat{f}_{h}\left ( x_{1},\cdots ,x_{n} \right )>k\textup{th percentile} \right \}.
\end{equation}
In this study, we established the threshold by considering the density values of all the points and selecting the value corresponding to the $k$th percentile.

Next, we utilize the Density-Based Spatial Clustering of Applications with Noise (DBSCAN) algorithm, which is a density-based spatial clustering method, to detect the main cluster and eliminate outlier clusters.
DBSCAN is able to find clusters with arbitrary shapes and clusters that contain noise, which is appropriate for processing the spatial data in this study.
The bounding polygon of the main cluster is then extracted to form a contour. When a vessel enters this contour, it is recognized as arriving at the PBG.

\begin{breakablealgorithm}
    \caption{Vessel arrival contour extraction}
    \label{vace}
    \begin{algorithmic}[1]
    \Input vessels' historical trajectories to a PBG
    \Output vessel arrival contour
        \State Scan histrical trajectories and get $X$ based on Eq. (\ref{eq_1});
        \State Apply MKDE to estimate the probability density function $f\left ( x_{1},x_{2} \right )$ with Eq. (\ref{eq_3});
        \State Get points $\hat{X}$ with high density using Eq. (\ref{eq_4});
        \State Apply DBSCAN to detect the main cluster and eliminate outlier clusters;
        \State Extract contour based on the main cluster.
    \end{algorithmic}
\end{breakablealgorithm}

\subsection{TCN-Based Model Using Multi-Data Fusion}
\label{tcn}
\begin{figure}[htbp]
\begin{center}
\includegraphics[width=0.43\textwidth]{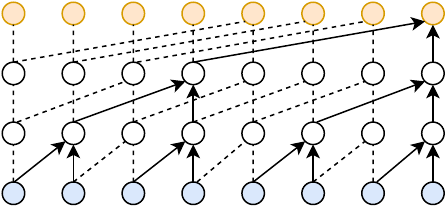}
\caption{Illustration of the dilated casual convolution.}
\label{dilated_conv}
\end{center}
\end{figure}

TCN is being used in this context due to its specific attributes. The notable components of TCN consist primarily of causal convolution, dilated convolution, and residual block \cite{zhang2023cost}.
Causal convolution is a specific type of convolution that is utilized in modelling temporal data to ensure that the temporal ordering is preserved. This means that when considering the output at time $t$, the convolution process can only involve elements from time $t$ or earlier in the previous layer. It cannot take into account any future time points beyond $t$. The purpose of using causal convolution in TCN is to prevent any leakage of information from the future to the past \cite{deng2022multi}, as shown by Fig. \ref{dilated_conv}.

Dilated convolution is a type of convolution where the kernel is expanded by creating holes between its elements. This expansion allows for a significant increase in the receptive field without sacrificing coverage \cite{yu2015multi}. The filters for dilated convolution can be represented by a function $F$ with a size of $f$, while $X$ represents the input time-series data. The convolved value at timestamp $t$ is computed using the following equation:
\begin{equation}\label{eq_dilated}
\textbf{v}(t)=(X\ast_{d}F)(t)=\sum_{i=0}^{f-1}F(i)\cdot X_{t-d\cdot i},
\end{equation}
where $d$ is the dilation rate and $f$ is the filter size. This equation essentially means that the value at time $t$ is obtained by multiplying the filter elements with the input data at time points $t-d\cdot i$ for $i=0,1,\cdots,f-1$ and summing the results.

\begin{figure}[htbp]
\begin{center}
\includegraphics[width=0.44\textwidth]{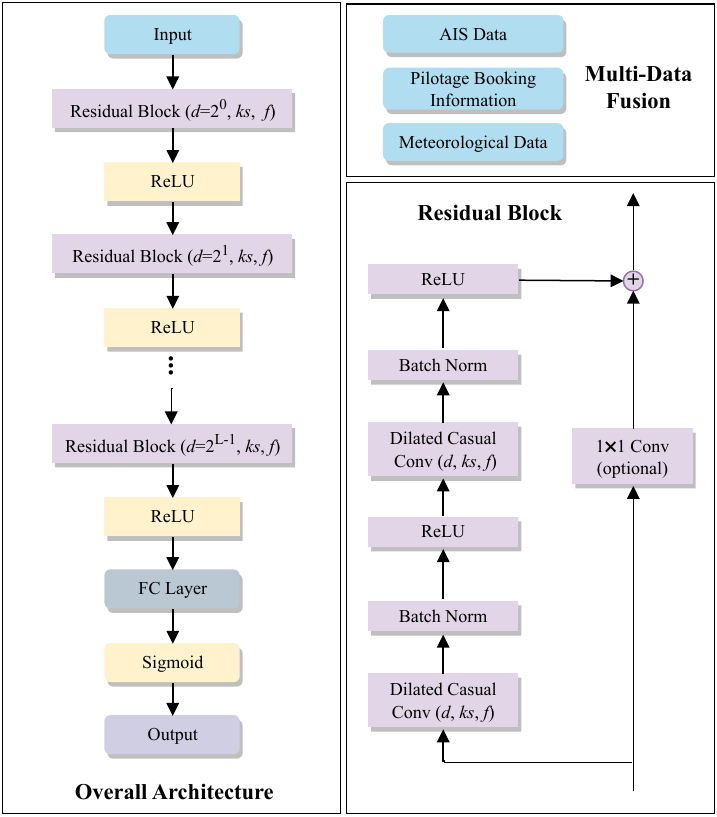}
\caption{The overall architecture of TCN for vessel arrival time prediction, where $d$ represents the dilation rate, $ks$ is the kernel size, $f$ stands for the filter size, and $L$ denotes the number of layers.}
\label{tcn_workflow}
\end{center}
\end{figure}

The architecture of the residual block, which is a crucial component, is shown in Fig. \ref{tcn_workflow}. In a single residual block, the input is processed through two stacked dilated causal convolutional layers and two Rectified Linear Unit (ReLU) activation layers. Additionally, two layers of batch normalization are used to speed up the network convergence and prevent overfitting. The output of the combined transformations is denoted by $\tau(X,W)$, where $X$ is the input and $W$ represents the weights. If the dimensions of the input and output are the same, the final output ($y$) of the residual block is obtained by adding the output of the transformation $\tau(X,W)$ to the input $X$, as shown by
\begin{equation}\label{eq_block_y}
y=\tau(X,W)+X.
\end{equation}
However, if the dimensions are different, a $1\times 1$ convolution is first applied to the input $X$ before the addition operation.

Assuming that we have the historical features of a vessel at timestamps $(t+1,t+2,\cdots,t+m)$, the objective is to predict when the vessel will arrive at the pilotage area.
The input $X$ of the model is represented by
\begin{equation}\label{eq7}
\begin{split}
\mathbf{x}_{j}=(( \varphi_{j1},\lambda_{j1},v_{j1},c_{j1},t_{j1},r_{j1},w^{s}_{j1},w^{d}_{j1}), \cdots ,\\ (\varphi_{jm},\lambda_{jm},v_{jm},c_{jm},t_{jm},r_{jm},w^{s}_{jm},w^{d}_{jm}))^{T},
\end{split}
\end{equation}
and
\begin{equation}\label{eq8}
\mathbf{X}=\left ( \mathbf{x}_{1},\mathbf{x}_{2},\cdot \cdot \cdot ,\mathbf{x}_{N-1},\mathbf{x}_{N}  \right ) \in \mathbb{R}^{N\times m\times 9},
\end{equation}
where $\varphi$, $\lambda$, $v$, $c$, $t$, $r$, $w^{s}$ and $w^{d}$ stands for the latitude, longitude, speed, course, time left until the Confirmed Service Time (CST), rainfall, wind speed and wind direction.
CST is defined as the confirmed service start time for the provision of pilotage service \cite{WinNT}.
The sequence length is denoted by $m$, and the number of observations is denoted by $N$. The rainfall and wind speed features are transformed into discrete binary features using an embedding process, as demonstrated by
\begin{equation}\label{eq_closs_1}
r_{ji}= \left\{
\begin{array}{lr} 
      1 & r_{ji}\geqslant  \varepsilon_{r}  \\
      0 & r_{ji}< \varepsilon_{r},
\end{array}
\right.
\end{equation}
and
\begin{equation}\label{eq_closs_2}
w^{s}_{ji}= \left\{
\begin{array}{lr} 
      1 & w^{s}_{ji}\geqslant  \varepsilon_{w^{s}}  \\
      0 & w^{s}_{ji}< \varepsilon_{w^{s}},
\end{array}
\right.
\end{equation}
where $\varepsilon _{r}$ and $\varepsilon _{w^{s}}$ are the thresholds for rainfall and wind speed, respectively. Furthermore, the wind direction is embedded using the Canonical Encoding Rules, as formulated by
\begin{equation}\label{eq_closs_3}
w^{d}_{ji}=\left ( \frac{\textup{sin}\left ( \theta_{ji}  \right )+1}{2}, \frac{\textup{cos}\left ( \theta_{ji}  \right )+1}{2} \right ).
\end{equation}

The ultimate output of the entire network can be obtained using Eq. (\ref{eq_block_y}), and a Fully-Connected (FC) layer and a sigmoid activation layer are attached to generate the output within the range of $[0, 1]$, as indicated by
\begin{equation}\label{eq9}
y=\textup{FC}(y),
\end{equation}
and
\begin{equation}\label{eq10}
y=\textup{sigmoid}(y).
\end{equation}

The groundtruth arrival time is denoted by $\hat{y}_{i}$, and the output from TCN is denoted by $y_{i}$. Therefore, the loss can be computed as follows:
\begin{equation}\label{eq_loss}
\mathcal{L}(W,b)=\frac{1}{N}\sum_{i=1}^{N}\left (y_{i}-\hat{y}_{i}  \right )^{2},
\end{equation}
where $N$ means the sample size, and $W$ and $b$ stand for the neural network's weights and biases, respectively.

During network training, the objective is to minimize the overall loss function by adjusting the weights and biases using the training data. This allows the network to learn and improve its predictions over time, as shown by
\begin{equation}\label{eq_minloss}
(W^{\ast },b^{\ast }) = \mathop{\arg \min}\limits_{W,b} \mathcal{L}(W,b),
\end{equation}
where $\mathcal{L}(W,b)$ is the overall loss function of Eq. (\ref{eq_loss}). To find the optimal parameters $(W^{\ast},b^{\ast})$, the widely-used back-propagation algorithm is employed in a similar way to train a basic deep neural network. An Adam optimizer is used during the back-propagation process.

\section{EXPERIMENTS AND EVALUATION}
\subsection{Data Sets}

\begin{table}[htbp]
\centering
\caption{Dataset Description}
\label{data_set}
\begin{threeparttable}
\begin{tabular}{|>{\centering}m{1.15cm}|>{\centering}m{1.2cm}|>{\centering}m{1.2cm}|>{\centering}m{1.2cm}|>{\centering}m{1.35cm}|}
\hline
PBG & Trajectory & Training Sample & Test Sample & Pred. Start Time (h)\\
\hline
PEBGA & 3,658 & 171,798 & 42,950 & 1.14\\
\hline
PEBGC & 8,901 & 334,716 & 83,680 & 1.30\\
\hline
\end{tabular}
\begin{tablenotes}
        \item[]\textit{Notes:} ``Pred. Start Time`` refers to the average time it takes for vessels to travel from the area where the predictions begin to the PBG.
\end{tablenotes}
\end{threeparttable}
\end{table}

Two real-world data sets, including AIS, pilotage booking data, and local meteorological data, from the Singapore Strait are used as experimental data in this study. The data spans 1 year, from January 1, 2018 to December 31, 2018. These data sets collect the trajectories of vessels heading to two PBGs in the Port of Singapore: the Eastern Boarding Ground “A” (PEBGA) and the Eastern Boarding Ground “C” (PEBGC). More details about the PBGs can be found at the PSA Marine's website\footnote[1]{\url{https://www.psamarine.com/wp-content/uploads/2017/08/Information-for-Merchant-Ships.pdf}}. The trajectories of vessels heading to these PBGs are extracted to construct the training and test data, with a preprocessing interpolation time interval of 1 minute. The earlier 80\% data are used for model training, and the later 20\% are used for tests. This division is made because the data follows a time-series pattern, and it is necessary to use only the preceding data for training \cite{zhang2019prediction}.
The predictions are consistently generated at intervals of 1 minute. Consequently, a single trajectory can produce multiple samples for both training and testing purposes.
It takes approximately 1.14-1.30 hours of travel time to reach the PBG from the prediction starting point.
More information about the data sets can be referred to Table \ref{data_set}.

\subsection{Experimental Settings}
Table \ref{para_set} shows the specific values used for each parameter.
For the extraction of the arrival contour, the value for $k$ (Eq. (\ref{eq_4})) was set to 75.
In the case of the TCN framework, we set the values of $m$ (Eq. (\ref{eq7})), $ks$ (Fig. \ref{tcn_workflow}), $f$ (Eq. (\ref{eq_dilated})) and $L$ (Fig. \ref{tcn_workflow}) to be 10, 15, 5, and 6 respectively.
For feature embedding, we used values of 7.6 mm and 22 knots for $\varepsilon _{r}$ (Eq. (\ref{eq_closs_1})) and $\varepsilon _{w^{s}}$ (Eq. (\ref{eq_closs_2})) respectively, which were scaled to represent heavy rain and strong breeze.
The performance of the models was assessed using the Mean Absolute Error (MAE), Root Mean Square Error (RMSE), and Coefficient of Determination ($\textup{R}^{2}$).
Python and TensorFlow were used to develop the models in this study.
\begin{table}[htbp]
\centering
\caption{Parameter Setting}
\label{para_set}
\begin{tabular}{|>{\centering}m{1.35cm}|>{\centering}m{1.2cm}|>{\centering}m{1.2cm}|>{\centering}m{1.2cm}|>{\centering}m{1.2cm}|}
\hline
Parameter & $k$ & $h$ & $m$ & $\varepsilon_{r}$ (mm)\\
\hline
Value & 75 & 0.001 & 10 & 7.6\\
\hline
Parameter & $\varepsilon_{w^{s}}$ (knot) & $ks$ & $f$ & $L$\\
\hline
Value & 22 & 15 & 5 & 6\\
\hline
\end{tabular}
\end{table}

\begin{figure}[htbp]
\begin{center}
\subfloat[]{
\includegraphics[width=0.1504\textwidth]{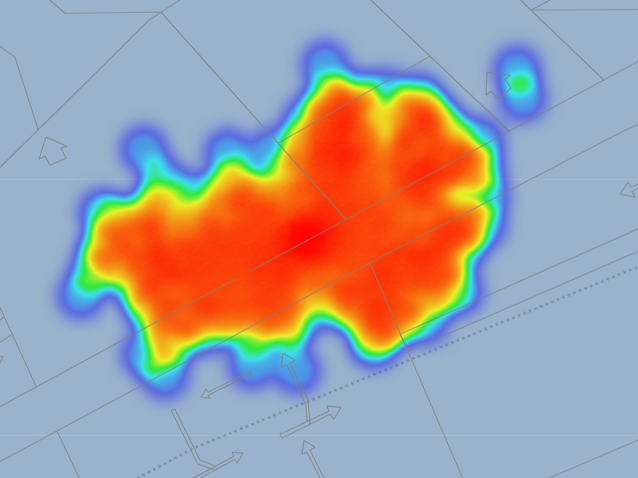}
}
\subfloat[]{
 \includegraphics[width=0.1504\textwidth]{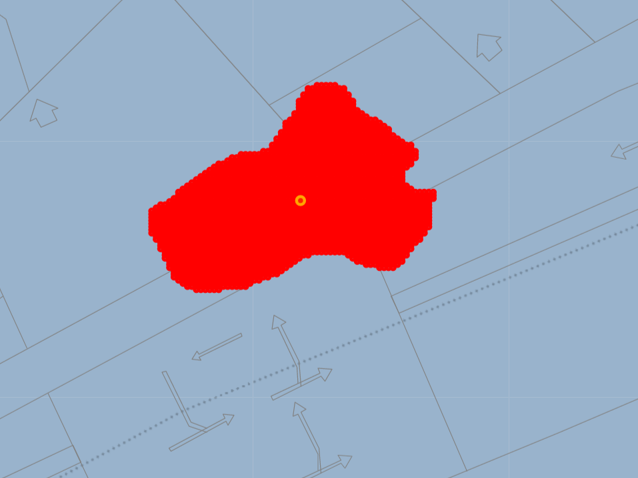}
 }
 \subfloat[]{
 \includegraphics[width=0.1504\textwidth]{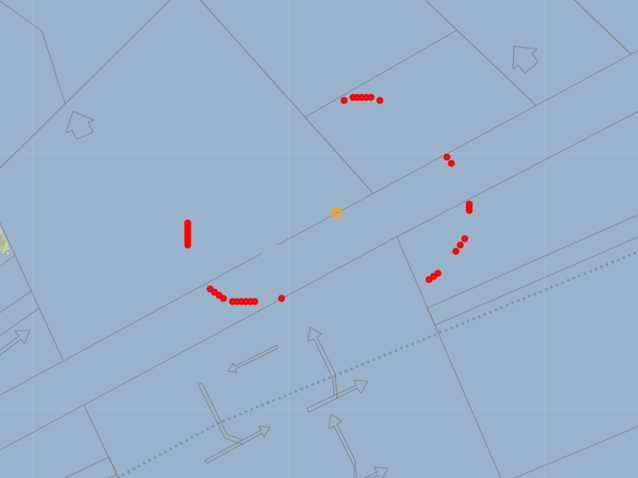}
 }
\caption{Vessel arrival contour extraction for PEBGA. (a) heatmap of the closest points to PEBGA. (b) refined area after clustering and removing outlier clusters. (c) extracted contour (bounding polygon).}
\label{fig_pebga}
\end{center}
\end{figure}
\begin{figure}[htbp]
\begin{center}
\subfloat[]{
\includegraphics[width=0.1504\textwidth]{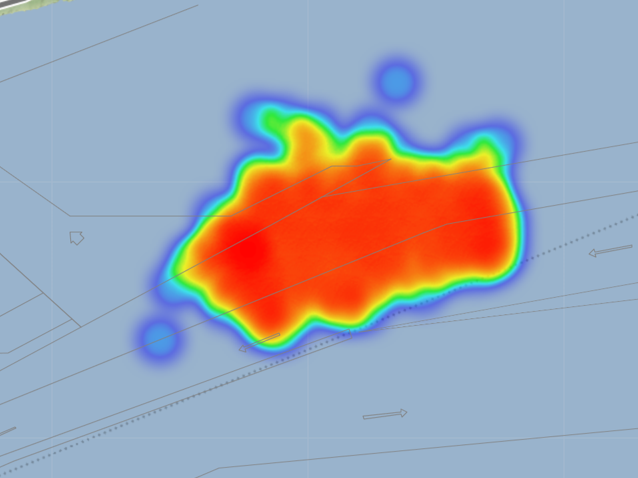}
}
\subfloat[]{
 \includegraphics[width=0.1504\textwidth]{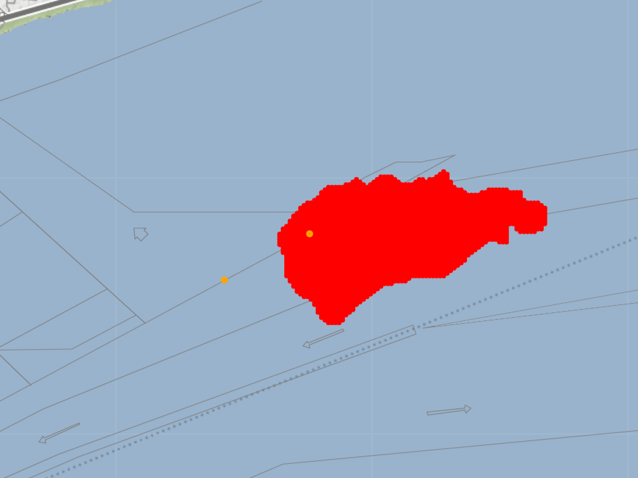}
 }
 \subfloat[]{
 \includegraphics[width=0.1504\textwidth]{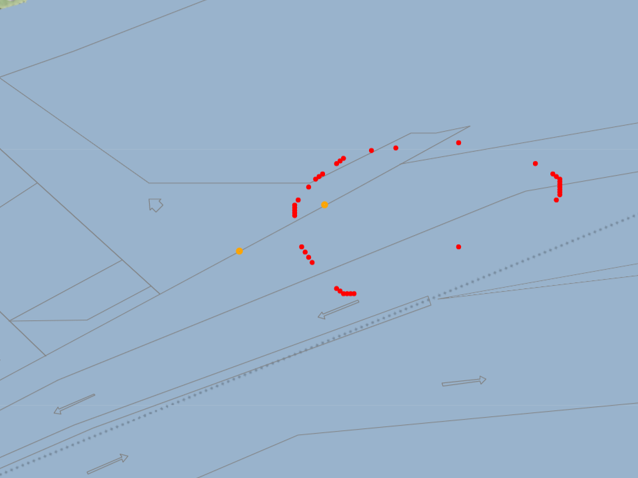}
 }
\caption{Vessel arrival contour extraction for PEBGC. (a) heatmap of the closest points to PEBGC. (b) refined area after clustering and removing outlier clusters. (c) extracted contour (bounding polygon).}
\label{fig_pebgc}
\end{center}
\end{figure}

\subsection{Results and Analyses}
Fig. \ref{fig_pebga} and Fig. \ref{fig_pebgc} depict the processes involved in obtaining the arrival contours for PEBGA and PEBGC.
Fig. \ref{fig_pebga}(a) and Fig. \ref{fig_pebgc}(a) show a heatmap that displays the closest points to the pilot boarding ground.
Fig. \ref{fig_pebga}(c) and Fig. \ref{fig_pebgc}(c) depict the refined area after applying MKDE and DBSCAN clustering.
The final arrival contours for PEBGA and PEBGC are shown in Fig. \ref{fig_pebga}(b) and Fig. \ref{fig_pebgc}(b), respectively.
These contours are utilized to define the arrival of the vessel and to determine the groundtruth arrival time for each trajectory.

\begin{table}[htbp]
\centering
\caption{Prediction Performance for PEBGA}
\label{res_pebga}
\begin{threeparttable}
\begin{tabular}{|>{\centering}m{2.7cm}|>{\centering}m{1.35cm}|>{\centering}m{1.5cm}|>{\centering}m{1.15cm}|}
\hline
Model & MAE (min) & RMSE (min) & $\textup{R}^{2}$\\
\hline
w/o rainfall \& wind & 4.66 & 7.99 & 0.8790\\
\hline
w/o CST & 6.37 & 9.87 & 0.8154\\
\hline
wth rainfall \& wind (continuous) & 4.75 & 7.41 & 0.8960\\
\hline
\textbf{wth rainfall \& wind (discrete)} & \textbf{4.58} & \textbf{6.82} & \textbf{0.9117}\\
\hline
\end{tabular}
\begin{tablenotes}
        \item[]\textit{Notes:} "w/o" and "wth" are short for "without" and "with", respectively.
\end{tablenotes}
\end{threeparttable}
\end{table}

\begin{table}[htbp]
\centering
\caption{Prediction Performance for PEBGC}
\label{res_pebgc}
\begin{threeparttable}
\begin{tabular}{|>{\centering}m{2.7cm}|>{\centering}m{1.35cm}|>{\centering}m{1.5cm}|>{\centering}m{1.15cm}|}
\hline
Model & MAE (min) & RMSE (min) & $\textup{R}^{2}$\\
\hline
w/o rainfall \& wind & 4.97 & 6.83 & 0.8813\\
\hline
w/o CST & 6.08 & 8.96 & 0.7981\\
\hline
wth rainfall \& wind (continuous) & 5.03 & 6.79 & 0.8828\\
\hline
\textbf{wth rainfall \& wind (discrete)} & \textbf{4.86} & \textbf{6.61} & \textbf{0.8890}\\
\hline
\end{tabular}
\begin{tablenotes}
        \item[]\textit{Notes:} "w/o" and "wth" are short for "without" and "with", respectively.
\end{tablenotes}
\end{threeparttable}
\end{table}

The prediction results of our method for PEBGA and PEBGC are presented in Table \ref{res_pebga} and Table \ref{res_pebgc}, respectively.
Incorporating meteorological data improves the $\textup{R}^{2}$ from 0.8790 to 0.9117 for PEBGA and from 0.8813 to 0.8890 for PEBGC.
However, pilotage information (i.e., CST) contributes more to the improvement of $\textup{R}^{2}$, enhancing it by 9.63\% (0.9117 vs. 0.8154) for PEBGA and by 9.09\% (0.8890 vs. 0.7981) for PEBGC.
Additionally, using discrete vectors to embed the wind and rainfall features improves the prediction performance for both data sets.
Specifically, compared to using continuous features, it reduces the MAE from 4.75 min to 4.58 min for PEBGA and from 5.03 min to 4.86 min for PEBGC.

\begin{figure}[htbp]
\begin{center}
\subfloat[PEBGA]{
    \includegraphics[width=0.231\textwidth]{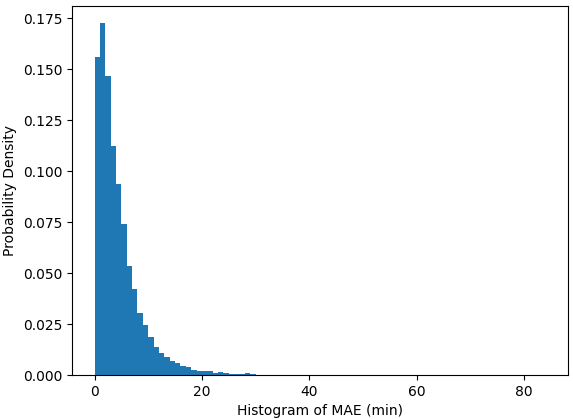}
    }
\subfloat[PEBGC]{
    \includegraphics[width=0.231\textwidth]{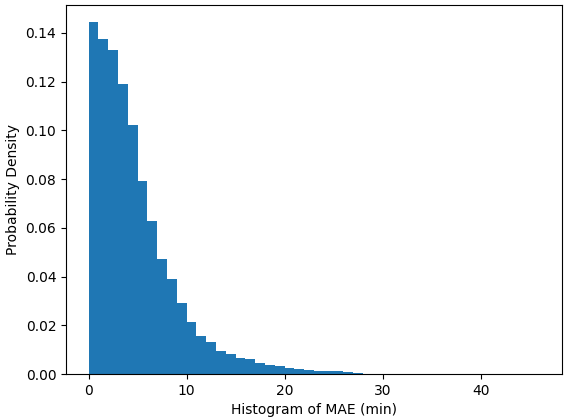}
    }
\caption{Histogram of MAE on the test data.}
\label{mae_hist}
\end{center}
\end{figure}

The proposed TCN framework is benchmarked with eight commonly used baseline models for time-series regression, including LSTM \cite{el2023deep}, GRU, Bidirectional LSTM (Bi-LSTM), Bidirectional GRU (Bi-GRU), Convolutional Neural Network (CNN) \cite{geng2019cost}, LSTM\_BiLSTM \cite{ma2021short}, LSTM Seq2Seq \cite{forti2020prediction}, and Stacked Bi-GRU (SBi-GRU) \cite{xu2022improved}.
The prediction results of TCN as well as the baseline models for PEBGA and PEBGC have been presented in Tables \ref{benchmark_pebga} and \ref{benchmark_pebgc}, respectively.
Our TCN model shows superior accuracy, with an MAE of only 4.58 min for PEBGA and 4.86 min for PEBGC, outperforming the best baseline model by at least 0.96\% (0.9117 vs. 0.9021 for PEBGA) in terms of $\textup{R}^{2}$. LSTM, GRU, Bi-LSTM, and Bi-GRU are popular RNN-based frameworks for sequential data modelling because of their strong capability of learning long-term interdependencies \cite{zhang2022arde}. Bi-LSTM and Bi-GRU can learn the sequential input data in both forward and backward directions, enabling a higher learning capability on sequential data. Fig. \ref{mae_hist}(a) and Fig. \ref{mae_hist}(b) show the histograms of MAE for PEBGA and PEBGC, respectively. For PEBGA, 90.61\% of the predictions' MAEs are within [0, 10 min], and 68.09\% are within [0, 5 min]. For PEBGC, 89.41\% of the predictions' MAEs are within [0, 10 min], and 63.65\% are within [0, 5 min].

\begin{table}[htbp]
\centering
\caption{Banchmarking of Prediction Performance for PEBGA}
\label{benchmark_pebga}
\begin{tabular}{|>{\centering}m{2.3cm}|>{\centering}m{1.4cm}|>{\centering}m{1.5cm}|>{\centering}m{1.3cm}|}
\hline
Model & MAE (min) & RMSE (min) & $\textup{R}^{2}$\\
\hline
LSTM & 4.74 & 7.34 & 0.8978\\
\hline
GRU & 4.76 & 7.37 & 0.8971\\
\hline
Bi-LSTM & 4.79 & 7.35 & 0.8976\\
\hline
Bi-GRU & 4.67 & 7.18 & 0.9021\\
\hline
CNN & 5.16 & 7.70 & 0.8876\\
\hline
LSTM\_BiLSTM & 4.92 & 7.50 & 0.8934\\
\hline
LSTM seq2seq & 5.60 & 8.21 & 0.8723\\
\hline
SBi-GRU & 5.05 & 7.62 & 0.8898\\
\hline
\textbf{TCN} & \textbf{4.58} & \textbf{6.82} & \textbf{0.9117}\\
\hline
\end{tabular}
\end{table}

\begin{table}[htbp]
\centering
\caption{Banchmarking of Prediction Performance for PEBGC}
\label{benchmark_pebgc}
\begin{tabular}{|>{\centering}m{2.3cm}|>{\centering}m{1.4cm}|>{\centering}m{1.5cm}|>{\centering}m{1.3cm}|}
\hline
Model & MAE (min) & RMSE (min) & $\textup{R}^{2}$\\
\hline
LSTM & 5.07 & 7.04 & 0.8739\\
\hline
GRU & 5.13 & 7.16 & 0.8698\\
\hline
Bi-LSTM & 5.08 & 7.08 & 0.8724\\
\hline
Bi-GRU & 5.08 & 6.98 & 0.8760\\
\hline
CNN & 5.64 & 7.98 & 0.8383\\
\hline
LSTM\_BiLSTM & 5.09 & 7.11 & 0.8714\\
\hline
LSTM seq2seq & 5.55 & 7.74 & 0.8478\\
\hline
SBi-GRU & 5.13 & 7.14 & 0.8703\\
\hline
\textbf{TCN} & \textbf{4.86} & \textbf{6.61} & \textbf{0.8890}\\
\hline
\end{tabular}
\end{table}

\section{CONCLUSIONS}
This paper proposes a method to predict vessel arrival time to pilot boarding grounds based on multi-data fusion and deep learning.
Initially, vessel arrival is defined using MKDE and DBSCAN clustering, which helps in extracting the vessel arrival contour. Subsequently, deep learning is applied using multi-data fusion, which incorporates AIS data, pilotage booking information, and meteorological data as inputs. A multi-layer TCN framework is utilized to recognize the vessel arrival pattern and predict the arrival time. The experiments carried out in Singapore, using two real-world data sets, demonstrate that the TCN framework for arrival time prediction is highly effective, achieving a high R2 value of 91.17\% and 88.90\%, with about 89.41\%-90.61\% of the absolute prediction residuals located within 10 minutes. Additionally, the proposed method outperforms other models such as LSTM, GRU, Bi-LSTM, Bi-GRU, CNN, LSTM\_BiLSTM, LSTM seq2seq, and SBi-GRU. The inclusion of pilotage booking information and meteorological data has been found to enhance prediction accuracy, with pilotage booking information being the more significant contributor. The use of discrete embedding for meteorological data is more effective than using continuous embedding.

\section{ACKNOWLEDGEMENT}
This study is partially supported by the Institute of High Performance Computing (IHPC)’s research project titled “Digital Intelligence Research Based on Big Data for Improving Efficient Navigation of Vessels to Singapore Port Waters” (Grant number SMI-2021-MTP-01 funded by Singapore Maritime Institute) and is also partially supported under Maritime AI Research Programme (Grant number SMI-2022-MTP-06 funded by Singapore Maritime Institute).


\bibliographystyle{IEEEtran}
\bibliography{IEEEabrv,Bibliography}


\end{document}